\documentclass[runningheads]{llncs}

\usepackage{eccv}
\usepackage{eccvabbrv}

\usepackage{graphicx}
\usepackage{booktabs}

\usepackage{amsmath}
\usepackage{amssymb}
\usepackage{multirow}
\usepackage{multicol}
\usepackage{makecell}
\usepackage{pifont}
\usepackage{array}
\usepackage{enumerate}
\usepackage{enumitem}
\usepackage{color}
\usepackage{tabularx}
\usepackage{pifont}
\usepackage{algorithm}
\usepackage{algpseudocode}
\usepackage{amsmath}
\usepackage{adjustbox}
\usepackage[dvipsnames,table]{xcolor} 
\usepackage{gensymb}

\usepackage[accsupp]{axessibility}  

\usepackage{hyperref}
\usepackage{orcidlink}

\begin{document}

\title{X-Imitator: Spatial-Aware Imitation Learning via Bidirectional Action-Pose Interaction} 

\titlerunning{X-Imitator}

\author{Kai Xiong \and 
Hongjie Fang \and
Lixin Yang \and
Cewu Lu}

\authorrunning{K.~Xiong et al.}

\institute{Shanghai Jiao Tong University}

\maketitle

\begin{abstract}
Effectively handling the interplay between spatial perception and action generation remains a critical bottleneck in robotic manipulation. Existing methods typically treat spatial perception and action execution as decoupled or strictly unidirectional processes, fundamentally restricting a robot's ability to master complex manipulation tasks. To address this, we propose X-Imitator, a versatile dual-path framework that models spatial perception and action execution as a tightly coupled bidirectional loop. By reciprocally conditioning current pose predictions on past actions and vice versa, this framework enables continuous mutual refinement between spatial reasoning and action generation. 
This joint modeling exactly mimics human internal forward models. 
Designed as a modular architecture, the system can be seamlessly integrated into various visuomotor policies. Extensive experiments across 24 simulated and 3 real-world tasks demonstrate that our framework significantly outperforms both vanilla policies and prior methods utilizing explicit pose guidance. The code will be open sourced.

\end{abstract}

\section{Introduction}
\label{sec:intro}

Robotic manipulation, with applications ranging from industrial assembly to domestic assistance, has been revolutionized by data-driven approaches in recent years~\cite{black2024pi_0,pi05,gr00t}. Specifically, imitation learning~\cite{pomerleau1988alvinn,bain1995framework,schaal1999imitation,brohan2022rt,zhao2023ACT,black2024pi_0,wang2024rise} has emerged as a simple but effective paradigm, allowing robots to acquire complex skills from pre-collected expert demonstrations~\cite{rh20t,robomind,droid}. Despite recent advancements in vision-based imitation learning, enabling robots to perform complex, dynamic manipulation where the spatial relationship between the robot and the object constantly evolves remains a fundamental challenge~\cite{liu2023target,ha2021flingbot}. A critical bottleneck in mastering these tasks lies in how existing frameworks handle the interplay between \textit{spatial perception} and \textit{action generation}.

To enhance spatial awareness in imitation policies, a common approach is to augment visual observations with 3D modalities like depth or point clouds~\cite{gervet2023Act3d,ze2024DP3,wang2024rise,fu2025cordvip,zhang2025FALCON,sun2025geovla,rise2}. However, relying on implicit feature learning from raw geometry often falls short in extracting the precise, object-level relationships needed for complex manipulation tasks. Another line of research utilizes explicit spatial guidance, such as flows~\cite{xu2024Im2Flow2Act,noh20253dflow,gu2023RTtrajectory,chen2026history,atm,general_flow} or visual prompts~\cite{tracevla,vpt}. Yet, flows are fundamentally low-level pixel representations lacking object-centric semantics, and visual prompts are typically confined to the 2D image plane, restricting their spatial reasoning capabilities. Fundamentally, the most direct and explicit way to embody spatial perception is through the object's 6D pose. Driven by this perspective, recent techniques leverage object pose trajectories to foster spatial-aware policy learning. However, they typically treat pose either as intermediate action~\cite{su2025MBA,hsu2025spot,bharadhwaj2024track2act} or as auxiliary supervision~\cite{graspvla}. Despite their efficacy, both paradigms fail to establish a true reciprocal connection. Specifically, auxiliary supervision only implicitly shapes shared representations during training, lacking any explicit interaction between pose and action during inference. Conversely, treating pose as the intermediate action enforces a strictly sequential, one-way pipeline where pose simply precedes action generation. In either case, the architecture prevents action execution from reciprocally informing or updating the spatial perception.

This decoupled, or unidirectional nature contrasts sharply with human manipulation, where \textit{spatial perception and action execution form a tightly coupled, bidirectional loop}. The intuition is straightforward: just as our perception of the object's current pose dictates our next precise movement, our recent actions inherently act as an internal forward model to predict and update the object's spatial state, especially when visual feedback is momentarily occluded by our own hands. To emulate this natural and synergistic process, we propose \textbf{X-Imitator}, a dual-path framework that jointly models action sequences and object pose trajectories. The framework is built upon three design principles:

\begin{enumerate}
    \item At its core lies a \textbf{bidirectional action-pose interaction}: the previously generated action serves as an explicit conditioning context to predict the current pose, and vice versa. As illustrated in Fig.~\ref{fig:Framework}, \textit{the prefix ``X'' precisely symbolizes this reciprocal, cross-branch flow of information}, enabling mutual refinement between spatial reasoning and action execution.
    \item This joint modeling inherently \textbf{eliminates dependence on external pose estimators} during inference. Conventional spatial-aware policies \cite{hsu2025spot} often depend on off-the-shelf pose estimation pipelines~\cite{wen2024foundationpose} during inference, which are computationally heavy and highly brittle under visual occlusions. By internally deducing the object's spatial state conditioned on its own actions, X-Imitator mitigates cascading perception errors, faithfully mirroring how humans rely on internal forward models when vision is obscured.
    \item X-Imitator is designed as a \textbf{versatile}, \textbf{plug-and-play} framework. Its action branch can be effortlessly instantiated by diverse existing imitation paradigms. To demonstrate this broad applicability, we build concrete instantiations upon representative 2D and 3D visuomotor policies~\cite{ze2024DP3,zhao2023ACT,wang2024rise}.
\end{enumerate}

Extensive experiments across 24 simulated and 3 real-world tasks demonstrate that this tightly coupled design significantly outperforms both vanilla policies and prior pose-enhanced frameworks, confirming the superiority of our reciprocal interaction mechanism. Furthermore, ablation studies confirm the superiority of our interactive design over non-interactive and alternative mechanisms. The code and data will be released upon acceptance.

\section{Related Works}

\subsection{Spatial Perception for Robotic Manipulation}

\subsubsection{Geometric Representations.}
To bridge the gap between 2D pixel inputs and the 3D physical world, many works augment visual observations with depth inputs, enabling native 3D geometric perception. PerAct \cite{shridhar2023PerAct} creates a voxelized 3D volume and applies Perceiver-IO~\cite{perceiver} to predict discrete actions. Act3D \cite{gervet2023Act3d} extends this by learning 3D feature fields of arbitrary spatial resolution via recurrent coarse-to-fine ghost point sampling. To improve efficiency, RVT \cite{goyal2023Rvt} avoids voxelization by projecting 3D point clouds into virtual orthographic views. While DP3 \cite{ze2024DP3} (a 3D extension of DP \cite{chi2023DP}) employs an MLP encoder to encode sparse point clouds into a compact 3D representation that conditions the diffusion process, RISE~\cite{wang2024rise} and RISE-2\cite{rise2} adopt the sparse 3D convolutional encoder~\cite{minkowski} and a Transformer~\cite{transformer} to obtain a more powerful featurization.
More recently, methods have integrated explicit 3D encoders directly into Vision-Language-Action (VLA) architectures \cite{qu2025spatialvla,li2025pointvla,sun2025geovla,li20253ds,huang2025graphcot} and utilized 3D perception capability of visual foundation models \cite{lin2025evo,guo2025glad,shi2025spatialactor,zhang2025FALCON}. For example, GeoVLA \cite{sun2025geovla} and PointVLA \cite{li2025pointvla} introduce a point encoder alongside the VLM, fusing them at the action head without disrupting the VLM backbone. Base on a frozen geometry-aware vision transformer VGGT \cite{wang2025vggt}, GLaD \cite{guo2025glad} and Evo-0 \cite{lin2025evo} extract geometric features from RGB images, thereby enhancing the spatial understanding of VLA. 
These methods have enhanced spatial awareness by dedicated architectural designs which are orthogonal approaches to our X-Imitator framework.

\subsubsection{Flow Representations.}
Recognizing the inherently dynamic nature of manipulation, another line of research employs flow as an intermediate representation to facilitate action prediction. While some studies utilize 2D flow to capture motion trends on the image plane \cite{xu2024Im2Flow2Act,bharadhwaj2024track2act,liu2025spatial,zhong2025flowvla}, others further lift it to 3D point flow for more accurate spatial modeling \cite{noh20253dflow,chen2026history,yang2025PPI,chen2025g3flow,dharmarajan2025dream2flow}. 
Despite its flexibility, flow-based representations---whether at the pixel level or their corresponding 3D point flows---can be computationally expensive and susceptible to noise. 
To address these limitations, X-Imitator introduces object pose trajectory as a lightweight, geometry-aware 3D representation, combining the dynamic advantages of flow methods with the structural robustness of rigid-body pose estimation.

\subsubsection{6D Pose Representations.} As a compact and precise representation of object state, pose is considered an intermediate representation of robot action in one line of research. For example, Track2Act \cite{bharadhwaj2024track2act} predicts point flow to infer rigid transformations of objects, while SPOT \cite{hsu2025spot} utilizes FoundationPose \cite{wen2024foundationpose} to track the current object pose, which then conditions future pose diffusion. These methods rely on a motion planner to convert the pose into end-effector action. In contrast, MBA \cite{su2025MBA} cascades pose estimation and action generation, eliminating the need for an external pose estimator or motion planner during inference. GraphCoT-VLA \cite{huang2025graphcot} constructs a 3D
Pose-Object graph to capture topological relationships between robot joints and objects. In this graph, however, each node is characterized by a 3D position rather than a 6D pose. Different from these approaches, our X-Imitator establishes a bidirectional action-pose interaction, going beyond the common practice of using pose merely as a one-way intermediate guidance for action generation.

\subsubsection{Visual Prompting Representations.}
To avoid learning complex spatial features from scratch, some approaches leverage the implicit spatial reasoning of Vision-Language Models (VLMs) or explicit visual prompts to guide policy learning. For instance, RoboPoint \cite{yuan2024robopoint} trains a VLM to generate 2D action points as spatial affordances, which are then projected into 3D targets for motion planning. In contrast, InternVLA-M1 \cite{chen2025internvla} adopts a locate-then-manipulate paradigm: once trained to output spatial coordinates, the model relies solely on text prompts to activate spatial reasoning during deployment.
2D trajectories overlaid on images offer a middle ground between motion expressiveness and ease of use. Consequently, recent methods explore history-aware \cite{zheng2024tracevla,patratskiy2025STvla}, present-aware \cite{dai2025aimbot}, and hindsight-aware \cite{gu2023RTtrajectory} 2D trajectories projected onto RGB or depth images to enable precise and smooth action generation. However, these methods often treat spatial information as a static pre-processing step.
Unlike such approaches, X-Imitator models object pose trajectories as dynamic variables that co-evolve with robot actions. Our interaction mechanism ensures mutual consistency between predicted object states and robot actions at every timestep, rather than merely conditioning one on the other.

\subsection{Dual-path Architecture and Iterative Refinement}
Dual-path (or dual-stream) architectures have become a dominant paradigm for handling heterogeneous information and decomposing feature dependencies. Originating from Siamese networks used in metric learning and visual tracking \cite{melekhov2016siamese,zhang2019deeper}, this architecture has evolved from learning symmetric similarity to modeling asymmetric complementary information. For instance, the dual-stream diffusion head \cite{won2025dual} maintains separate streams for action and image modalities while enabling cross-modal knowledge sharing. And the dual-path point embedding network \cite{sun2025geovla} extracts a geometry representation with precise spatial location.

Besides robotic manipulations, dual-path designs are commonly employed for cross-modal alignment and interaction in multi-modal tasks \cite{radford2021clip,lu2022cots,liu2024duapin}, as well as for spatiotemporal decomposition in video analysis \cite{han2022dualAI,xiao2025d2Stream}. Among these applications, cross-attention serves as a standard mechanism for information exchange. Note that existing dual-path architectures typically perform information interaction within a single forward pass, whereas our X-Imitator facilitates interaction across inference steps. This temporal form of interaction resembles an iterative refinement module, which recursively takes its own output as input for the next step to enhance prediction precision \cite{teed2020raft,li2018deepim,wang2019densefusion}. Consequently, X-Imitator marries the merits from both sides: the dual-path structure achieves flexibility of our framework, while the across-step bidirectional interaction enables coevolution of action and pose predictions.

\section{Method}

In this section, we formally present X-Imitator, a spatial-aware imitation learning framework with a bidirectional action-pose interaction mechanism, designed to explicitly incorporate spatial information into modern visuomotor policies.

\subsection{Preliminary}

Let $\mathbf{O}_t$ denote the visual observation at time step $t$, which typically includes RGB-D images and robot proprioception. The objective is to learn a policy $\pi$ that predicts the future action trajectory $\mathbf{A}_t = \{a_t, a_{t+1}, \dots, a_{t+H-1}\}$ given $O_t$, where $H$ is the prediction horizon.
Unlike standard approaches that map $\mathbf{O}_t \rightarrow \mathbf{A}_t$ alone, X-Imitator simultaneously predicts the object pose trajectory $\mathbf{P}_t = \{p_t, p_{t+1}, \dots, p_{t+H-1}\}$, where $p_i \in SE(3)$, represented as a 9D vector containing 3D translation and 6D rotation \cite{zhou2019continuity}, is the pose of a target object in the camera or robot base coordinate system. We consider the objects or articulated object parts being manipulated as the target objects. With a little symbol abuse, $p_i \in \mathbb{R}^{9n}$ also represents the concatenated poses when there are $n > 1$ target objects.

\begin{figure}[tb]
\centering
\includegraphics[width=1.0\linewidth]{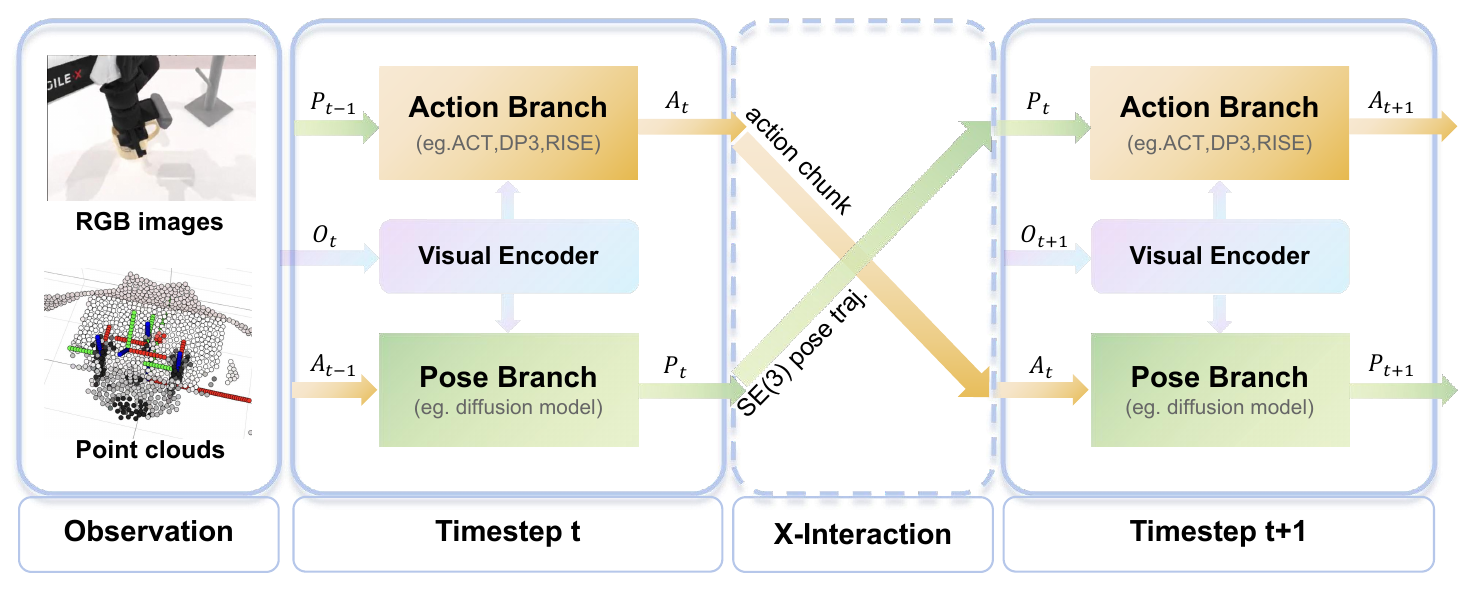}
\caption{\label{fig:Framework} Overview of X-Imitator. The framework maintains two interactive branches: an action branch (the policy) and a pose branch with shared visual features. The key innovation is an across-step interaction mechanism: the output from one branch at timestep $t$ serves as an additional conditional input for the other branch at timestep $t+1$.}
\end{figure}

\subsection{Architecture Design}

As shown in Figure \ref{fig:Framework}, the overall architecture of X-Imitator consists of two parallel branches: an action branch $\mathcal{F}_{a}$ and a pose branch $\mathcal{F}_{p}$. Both branches share a common visual encoder to extract perceptual features $F_t^{vis}$ from observation $\mathbf{O}_t$. The core innovation lies in the dual-branch action-pose interaction mechanism. 

\subsubsection{Bidirectional Action-Pose Interaction.} Instead of treating action and pose prediction as independent tasks, we introduce a bidirectional information flow between them. Specifically, the action branch predicts $\mathbf{A}_t$ conditioned on the visual feature $F_t^{vis}$ and the previous pose estimate $\mathbf{P}_{t-1}$. Conversely, the pose branch predicts $\mathbf{P}_t$ conditioned on $F_t^{vis}$ and the previous action estimate $\mathbf{A}_{t-1}$. This design forms a closed-loop interaction between spatial perception and motion execution, allowing the two branches to iteratively refine each other and yielding more informed, geometrically grounded policy decisions. Formally, the update rules of X-Imitator are defined as:
\begin{equation}\label{update}
\left\{
\begin{aligned}
    &\mathbf{A}_t = \mathcal{F}_{a}(\text{Cond}_a(\phi_a(\mathbf{P}_{t-1}), F_t^{vis})) \\
    &\mathbf{P}_t = \mathcal{F}_{p}(\text{Cond}_p(\phi_p(\mathbf{A}_{t-1}), F_t^{vis}))
\end{aligned}\right.
\end{equation}
where $\phi(\cdot)$ projects the past trajectory into feature space, and $\text{Cond}(\cdot)$ is a feature fusion operator which degrades to output $F_t^{vis}$ directly when there is no past estimate ($t=0$). 

\subsection{Instantiations}

\begin{figure}[tb]
\centering
\includegraphics[width=1.0\linewidth]{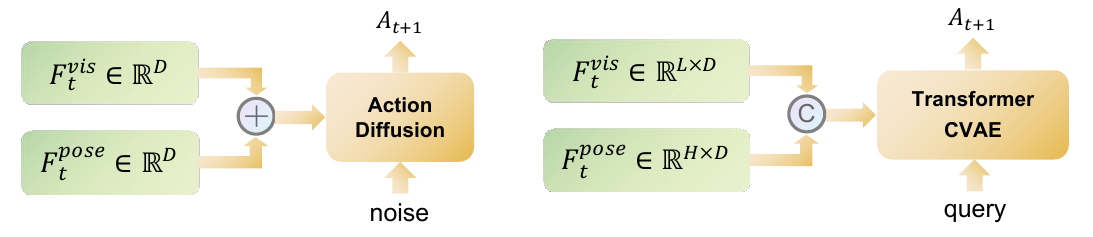}
\caption{\label{fig:Instantiation} Feature fusion in action branch. Left: Add fusion in X-DP3. Right: Concat fusion in X-ACT. The visual feature $F_t^{vis}$ is extracted by the perception module of each base method. The pose trajectory feature $F_t^{pose}$ is obtained by: MLP on flattened $P_{t-1}$ for X-DP3 and X-RISE; self-atten on $P_{t-1}$ + cross-atten with $F_t^{vis}$ for X-ACT.}
\end{figure}

To demonstrate the versatility of X-Imitator, we instantiate the action branch using three representative visuomotor policies: DP3\cite{ze2024DP3}, ACT\cite{zhao2023ACT}, and RISE\cite{wang2024rise}. For the pose branch, we implement it as a lightweight diffusion head \cite{ho2020DDPM,chi2023DP} for simplicity.  
While these instantiations share the high-level dual-path logic, the key differences lie in how the conditional sequence projector $\phi(\cdot)$ and feature fusion operator $\text{Cond}(\cdot)$ in Eqn.~\eqref{update} are implemented to suit the base architecture.
Fig.~\ref{fig:Instantiation} illustrates the feature fusion in action branch.

\subsubsection{X-DP3.}
DP3 uses a U-Net-based diffusion model operating on point cloud features $F_t^{vis}$. In X-DP3, the interaction is implemented via direct feature injection. Specifically, we employ a three-layer MLP to project each of the flattened sequences ($\mathbf{A}_{t-1}$, $\mathbf{P}_{t-1}$) into conditional features ($F_t^{act}$, $F_t^{pose}$), which are then fused with the visual features $F_t^{vis}$ via element-wise addition. The resulting combined features are fed into the respective diffusion heads for both action and pose prediction ($\mathbf{A}_t$, $\mathbf{P}_t$). Empirically, we find that the straightforward additive fusion outperforms the more delicate FiLM fusion in our case. A similar phenomenon is also observed in~\cite{zhang2025FALCON}. 

\subsubsection{X-ACT.}
ACT uses a transformer-based conditional variational auto-encoder to generate robot actions, with multi-view RGB images as visual inputs. For X-ACT, we also leverage the sequence modeling capabilities of the transformer. Each conditional sequence ($\mathbf{A}_{t-1}$, $\mathbf{P}_{t-1}$) is first linearly projected into a sequence of tokens. These tokens undergo self-attention to capture temporal dependencies. Crucially, we then apply cross-attention, where the condition tokens serve as queries and the ResNet-based visual tokens $F_t^{vis}$ serve as keys and values. This produces two sets of visually-enhanced condition tokens ($F_t^{act}$, $F_t^{pose}$). 
In the action branch, $F_t^{pose}$ is concatenated with $F_t^{vis}$ to feed into the transformer to generate action $\mathbf{A}_t$. In the pose branch, $F_t^{act}$ and $F_t^{vis}$ are reduced along sequence dimension, followed by element-wise addition, to obtain the fused feature vector to guide the pose trajectory $\mathbf{P}_t$ diffusion.

\subsubsection{X-RISE.}
RISE utilizes a 3D sparse encoder to extract point cloud tokens, which are aggregated into a feature vector $F_t^{vis}$ by a transformer. Specifically, the point tokens are processed by a transformer encoder first. Then, for each decoding step in the transformer, one readout token is utilized to query the processed point tokens to obtain $F_t^{vis}$. 
In X-RISE, similar to X-DP3, each conditional sequence ($\mathbf{A}_{t-1}$, $\mathbf{P}_{t-1}$) is flattened and projected via a linear layer to obtain conditional features ($F_t^{act}$, $F_t^{pose}$), which act as two readout tokens to query the processed point tokens, resulted in the visually-enhanced version of conditional features. Unlike X-ACT, we empirically find that these two features are sufficiently informative to guide the action and pose diffusion without further concatenating with $F_t^{vis}$. 

\subsubsection{Training Objective.}
The framework is trained end-to-end using a joint loss function defined as:
\begin{equation}
    \mathcal{L} = \mathcal{L}_\text{action} + \lambda \cdot \mathcal{L}_\text{pose},
    \label{loss}
\end{equation}

\begin{itemize}
    \item $\mathcal{L}_\text{action}$: For X-ACT, this is the standard CVAE reconstruction loss (L1 loss on actions + KL divergence). For X-DP3 and X-RISE, this is the diffusion denoising MSE loss on the action sequence.
    
    \item $\mathcal{L}_\text{pose}$: This is the diffusion denoising MSE loss on the pose trajectory. $\lambda$ is a hyperparameter balancing the two tasks, and we fix it to 1.
\end{itemize}

\section{Simulation Experiments} 
In this section, we empirically evaluate X-Imitator across diverse tasks in several simulation benchmarks. We aim to answer the following research questions:
\begin{enumerate}[leftmargin=1.5cm]
    \item[\textbf{(RQ1)}] Can X-Imitator improve the performance of various visuomotor policies by leveraging bidirectional action-pose interaction?
    \item[\textbf{(RQ2)}] Does the bidirectional action-pose interaction mechanism lead to better performance than other interactive or non-interactive alternatives?
    \item[\textbf{(RQ3)}] How does the conditional trajectory length in bidirectional action-pose interaction influence the policy performance?
\end{enumerate}

\subsection{Setup}

\begin{figure}[tb]
\centering
\includegraphics[width=0.9\linewidth]{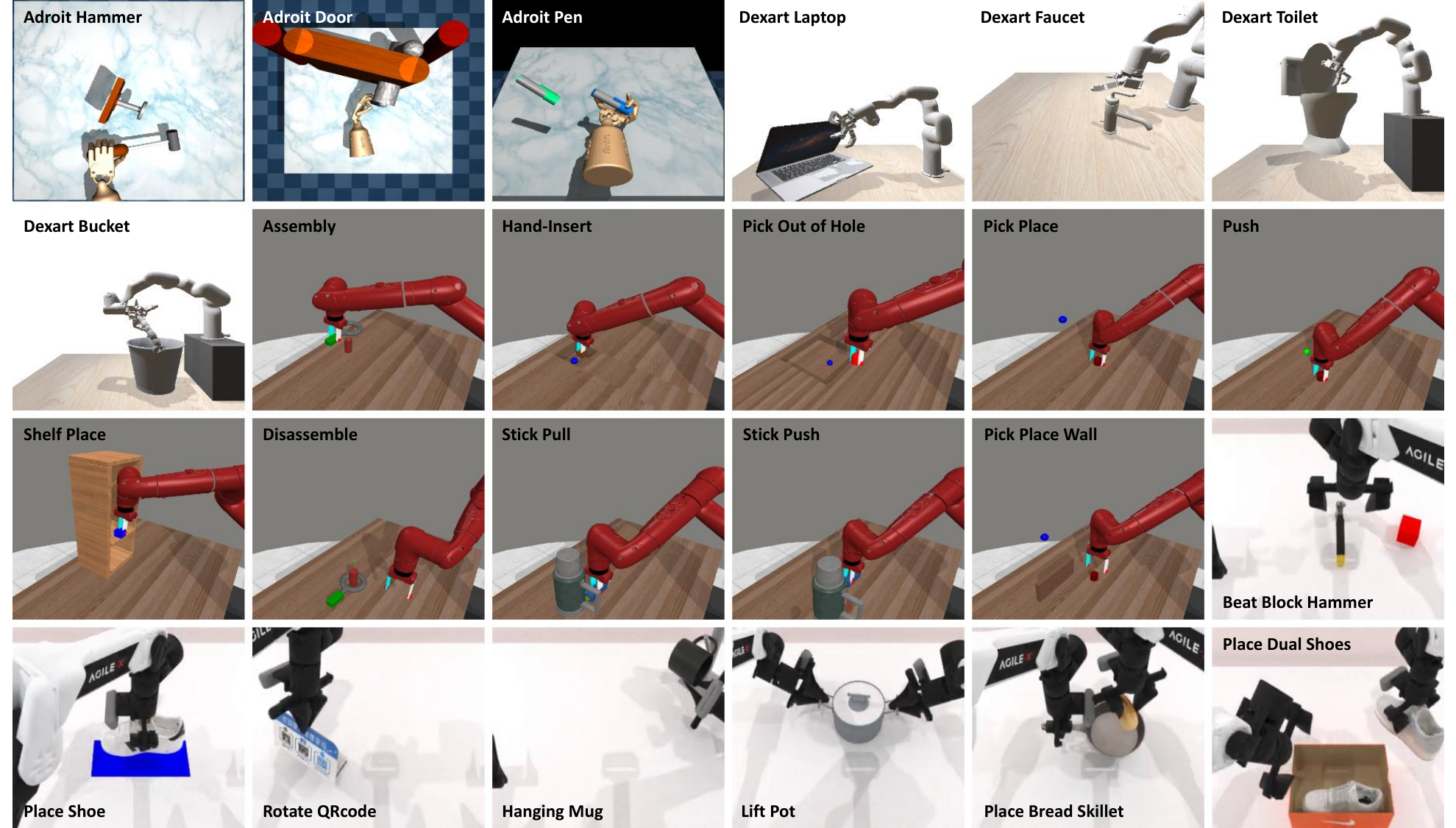}
\caption{\label{fig:sim_tasks} Overview of simulated tasks. The first 3 and 4 tasks are from Adroit and Dexart, respectively. The middle 10 tasks with a red robot arm are from MetaWorld, and the last 7 tasks with dual arms are from RoboTwin 2.0.}
\end{figure}

\subsubsection{Environments and Benchmarks.} To comprehensively evaluate our framework, we select a total of 24 simulation tasks across four diverse benchmarks, as illustrated in Fig.~\ref{fig:sim_tasks}. Following prior works \cite{ze2024DP3,su2025MBA,su2025dense}, our evaluation spans Adroit~\cite{rajeswaran2017learning}, DexArt~\cite{bao2023dexart}, MetaWorld~\cite{yu2020meta}, and the recent dual-arm RoboTwin 2.0~\cite{chen2025robotwin} simulation benchmarks. Specifically, we evaluate on the \textit{hard} and \textit{very hard} settings of MetaWorld (excluding the \textit{push back} task in \textit{hard} due to expert script failures as reported in~\cite{ze2024DP3}). For RoboTwin 2.0, we carefully select 7 tasks requiring precise spatial coordination, which encompass both simultaneous dual-arm manipulation and scene-dependent single-arm selection. Overall, these tasks involve rigid and articulated objects, single-arm and dual-arm configurations, as well as gripper and dexterous hand end-effectors, providing a rigorous testbed for spatial-aware manipulation.

\subsubsection{Baselines.}
To validate the effectiveness of the proposed framework, we employ the representative point-based policies DP3 \cite{ze2024DP3}, RISE \cite{wang2024rise}, and one image-based policy ACT \cite{zhao2023ACT}, as our baselines. 
DP3 with MBA enhancement~\cite{su2025MBA}. As detailed in the method section, our X-DP3, X-ACT, and X-RISE are developed based on these baselines to show improvements brought by our framework. We use the simple configuration for DP3 and adopt a smaller model for ACT. We fix bugs of the DP3 released code involving point coordination system misuse in Adroit and target missing issues in MetaWorld.

\subsubsection{Implementations.} We generate 10 episodes for Adroit and MetaWorld tasks, 50 episodes for RoboTwin 2.0 tasks, and 100 episodes for DexArt tasks for training. The policies are trained with a batch size of 128 and a learning rate of $10^{-4}$. The horizon $H$ is set to 48 for (X-)ACT and 16 for others. (X-)DP3 uses observations of both $O_t$ and $O_{t-1}$ while others use $O_t$ only. For our instantiations, we introduce an extra parameter $N$, defaulting to 8, which represents the conditional trajectory length. This design considers real scenarios where the policy predicts, for example, $H=16$ steps, but only executes $N=8$ steps before the next prediction. By decoupling horizon and conditional length, we can adjust the number of execution steps at the interval of $[N,H]$ during inference without requiring specific treatment.
To boost robustness and mitigate compounding error, we perform condition augmentation by adding noise to the ground-truth action/pose in the training stage, mimicking prediction error during inference. 

\subsubsection{Evaluation Protocols.} We evaluate all policies for 100 consecutive trials on the RoboTwin 2.0 benchmark and 300 trials on the other benchmarks after training. Only the final checkpoint is evaluated, and we record its average success rate as the metric. This is slightly different from the metric used in~\cite{ze2024DP3,su2025MBA}, where they evaluated the policy during training and computed the average of the highest 5 success rates. We choose to evaluate the final trained model since it is more realistic and closer to real-world practical settings.

\begin{table*}[t]
\caption{Simulation success rates (\%) on Adroit, DexArt and MetaWorld benchmarks. Detailed comparison for each task is reported in Fig.~\ref{fig:bar_charts}.}
\label{tab:simulation_ADM}
\centering
\footnotesize
\setlength{\tabcolsep}{4.5pt}
\begin{tabular}{lccccc}
\toprule
\textbf{Method} & \textbf{Adroit} & \textbf{DexArt} & \begin{tabular}{c}\textbf{MetaWorld} \\ \textit{(hard)}\end{tabular} & \begin{tabular}{c}\textbf{MetaWorld} \\ \textit{(very hard)} \end{tabular} & \textbf{Average} \\
\midrule
DP3~\cite{ze2024DP3} & 62.2 & 55.9 & 48.9 & 51.4 & 54.6\\
DP3+MBA~\cite{su2025MBA} & 63.2 & 57.0 & 52.8 & 62.7 & 58.9 \\ 
\midrule
DP3+UniAux & 64.0 & 59.3 & 49.7 & 52.1 & 56.3 \\
DP3+SepAux & 66.6 & 57.4 & 52.6 & 64.3 & 60.2 \\
DP3+Hybrid & 65.3 & 57.3 & 54.7 & 64.4 & 60.4 \\ 
\midrule
X-DP3 \textit{(ours)} & \cellcolor[HTML]{ecf8f8}\textbf{71.4} & \cellcolor[HTML]{ecf8f8}\textbf{59.4} & \cellcolor[HTML]{ecf8f8}\textbf{57.7} & \cellcolor[HTML]{ecf8f8}\textbf{66.6} & \cellcolor[HTML]{ecf8f8}\textbf{63.8} \\
\bottomrule
\end{tabular}
\end{table*}

\subsection{Results} 

\subsubsection{Establishing a tightly coupled action-pose interaction via X-Imitator consistently enhances the spatial awareness and overall performance across diverse visuomotor policies (RQ1).} 
\begin{figure}[tb]
\centering
\includegraphics[width=1.0\linewidth]{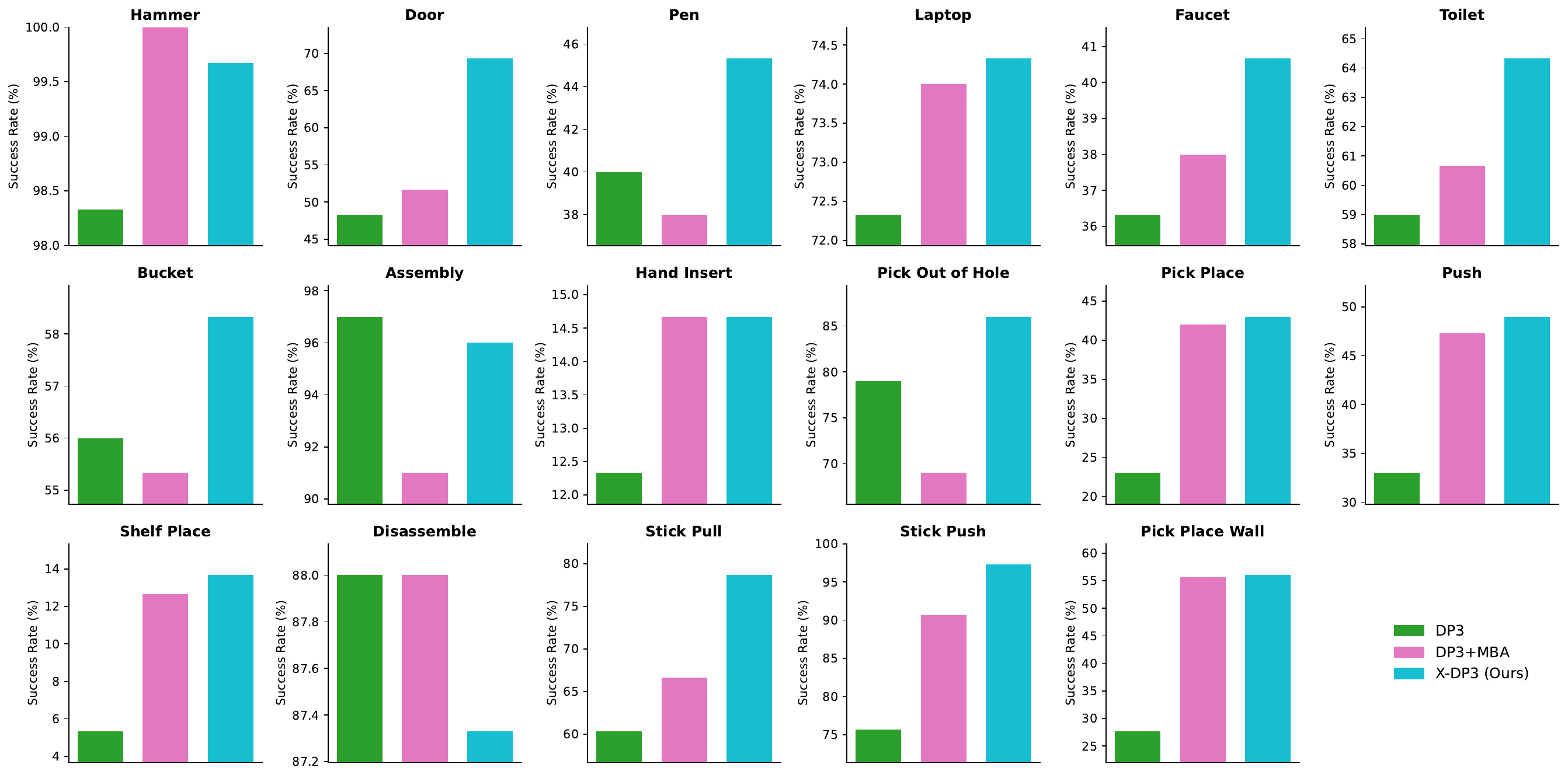}
\caption{\label{fig:bar_charts} Comparison of our X-DP3 against DP3 and its MBA-enhanced version, over each of the 17 simulated tasks from Adroit, Dexart and MetaWorld. The average success rate of each simulated benchmark is reported in Table \ref{tab:simulation_ADM}.}
\end{figure}

We demonstrate this through both interaction effectiveness and framework generality. Regarding effectiveness, X-DP3 achieves a 63.8\% average success rate across Adroit, DexArt, and MetaWorld (\textit{hard} and \textit{very hard}) benchmarks, outperforming the base DP3 (54.6\%) and DP3+MBA (58.9\%), as shown in Table~\ref{tab:simulation_ADM}. This performance gain, driven by the integration of utilizing pose trajectories and a bidirectional action-pose interaction mechanism, is especially evident in tasks requiring precise spatial alignment (e.g., \textit{Door}, \textit{Pen}, and \textit{Stick Pull}), where X-DP3 yields improvements up to 21\%, as demonstrated in Figure \ref{fig:bar_charts}. Furthermore, we show that X-Imitator exhibits strong generality across diverse base policies on the RoboTwin 2.0 benchmark in Table \ref{tab:simulation_robotwin}. Notably, it provides a substantial gain for the image-based ACT (27.86\% to 37.57\%), which inherently lacks spatial awareness due to its 2D input and thus benefits the most from the introduced pose interactions. Importantly, these consistent improvements span diverse manipulation scenarios, demonstrating the framework's robustness across various end-effectors and both single-arm and dual-arm configurations.

\subsubsection{Computational Overhead.}
As an example, compared to RISE with 51M parameters, X-RISE has 71M parameters which slightly increases the memory usage and training time per iteration from (4220M,228ms) to (4726M,243ms). During inference, due to the added pose diffusion head, X-RISE drops from 11Hz to 6Hz, predicting 6*Horizon steps per second.
Note that the two branches of X-RISE are implemented sequentially and could be accelerated via parallel execution.

\begin{table*}[tb]
\caption{Simulation success rates (\%) on RoboTwin  2.0 benchmark. The task names are simplified to ensure a clean presentation. Note that while tasks in the Single-Arm category can be performed with one arm, the policy must determine which arm to use based on the scene configuration.}
\label{tab:simulation_robotwin}
\centering
\footnotesize
\setlength{\tabcolsep}{5pt}
\begin{tabular}{lcccccccc}
\toprule
\multirow{2}{*}{\textbf{Method}} & \multicolumn{3}{c}{\textbf{Single-Arm}} & \multicolumn{4}{c}{\textbf{Dual-Arm}} & \multirow{2}{*}{\textbf{Average}} \\
\cmidrule(lr){2-4} \cmidrule(lr){5-8} 
 & hammer & shoe & QRcode & mug & pot & bread & shoes \\ \midrule
ACT~\cite{zhao2023ACT} & 55 & 7 & 28 & 10 & 82 & \cellcolor[HTML]{ecf8f8}\textbf{10} & 3 & 27.86 \\
X-ACT \textit{(ours)} & \cellcolor[HTML]{ecf8f8}\textbf{65} & \cellcolor[HTML]{ecf8f8}\textbf{31} & \cellcolor[HTML]{ecf8f8}\textbf{46} & \cellcolor[HTML]{ecf8f8}\textbf{18} & \cellcolor[HTML]{ecf8f8}\textbf{87} & \cellcolor[HTML]{ecf8f8}\textbf{10} & \cellcolor[HTML]{ecf8f8}\textbf{6} & \cellcolor[HTML]{ecf8f8}\textbf{37.57} \\
\midrule
DP3~\cite{ze2024DP3} & 81 & 52 & 64 & 23 & 92 & 38 & 18 & 52.57 \\
X-DP3 \textit{(ours)} & \cellcolor[HTML]{ecf8f8}\textbf{87} & \cellcolor[HTML]{ecf8f8}\textbf{59} & \cellcolor[HTML]{ecf8f8}\textbf{69} & \cellcolor[HTML]{ecf8f8}\textbf{27} & \cellcolor[HTML]{ecf8f8}\textbf{98} & \cellcolor[HTML]{ecf8f8}\textbf{48} & \cellcolor[HTML]{ecf8f8}\textbf{20} & \cellcolor[HTML]{ecf8f8}\textbf{58.29} \\
\midrule
RISE~\cite{wang2024rise} & 90 & \cellcolor[HTML]{ecf8f8}\textbf{78} & 59 & 39 & 91 & 44 & \cellcolor[HTML]{ecf8f8}\textbf{20} & 60.14 \\
X-RISE \textit{(ours)} & \cellcolor[HTML]{ecf8f8}\textbf{92} & 72 & \cellcolor[HTML]{ecf8f8}\textbf{63} & \cellcolor[HTML]{ecf8f8}\textbf{62} & \cellcolor[HTML]{ecf8f8}\textbf{93} & \cellcolor[HTML]{ecf8f8}\textbf{52} & 17 & \cellcolor[HTML]{ecf8f8}\textbf{64.43} \\
\bottomrule
\end{tabular}
\end{table*}

\subsection{Ablations} 
\subsubsection{The bidirectional action-pose interaction is more effective than the unidirectional interaction and non-interactive alternatives for capturing spatial dependencies (RQ2).}

\begin{figure}
\centering
\includegraphics[width=1.0\linewidth]{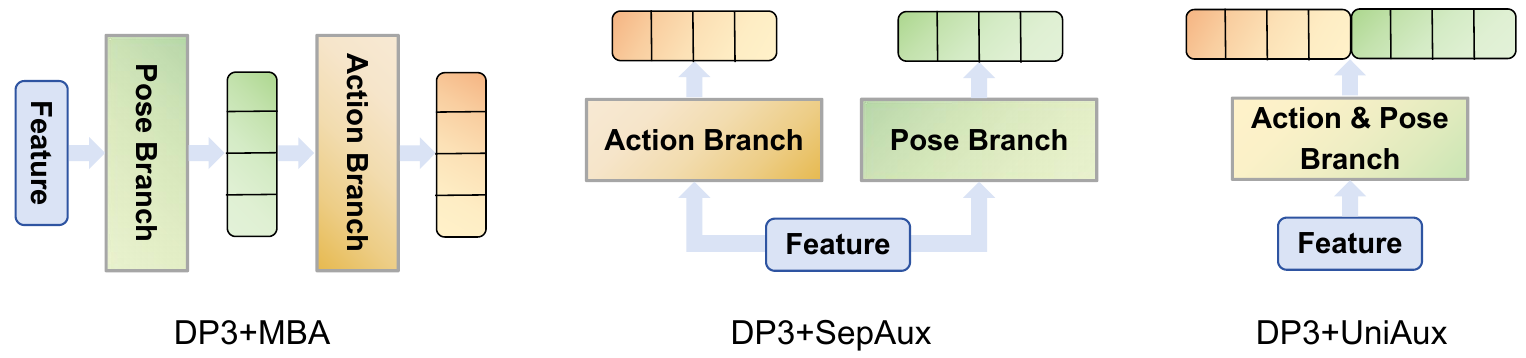}
\caption{\label{fig:ablations} Comparison of DP3 variants in the ablation studies.}
\end{figure}

We design three DP3 variants, with their differences illustrated in Fig.~\ref{fig:ablations}, which also includes DP3+MBA~\cite{su2025MBA} for a complete comparison.
DP3+SepAux adopts a multi-task learning paradigm without branch interaction, using pose prediction as an auxiliary task. DP3+UniAux takes the concatenated vector of action and pose as the unified diffusion target. DP3+Hybrid utilizes a hybrid training strategy that, in each iteration, the model is trained in X-DP3 style or in DP3-SepAux style, each with the probability of 0.5. We aim to make the action branch compatible with the presence or absence of the conditional pose trajectory. Therefore, the model runs two branches to learn task-related spatial information during training, while it only keeps the action branch during inference.
Results shown in Table \ref{tab:simulation_ADM} (middle rows) demonstrate that X-DP3 outperforms these variants, confirming that the dual-path bidirectional interaction is more effective than other alternatives for capturing spatial dependencies. 
We can see that DP3+Hybrid degrades to DP3+SepAux. The reason might be that its model capacity is insufficient to support mixed training. We can also observe that the simple UniAux strategy brings a slight improvement (1.7\%) over DP3, again showing the benefits of introducing bidirectional pose-action interaction for better spatial perception.

\subsubsection{A longer conditional trajectory can better capture motion patterns and tends to yield better performance (RQ3).}
\begin{figure}[tb]
\centering
\includegraphics[width=1.0\linewidth]{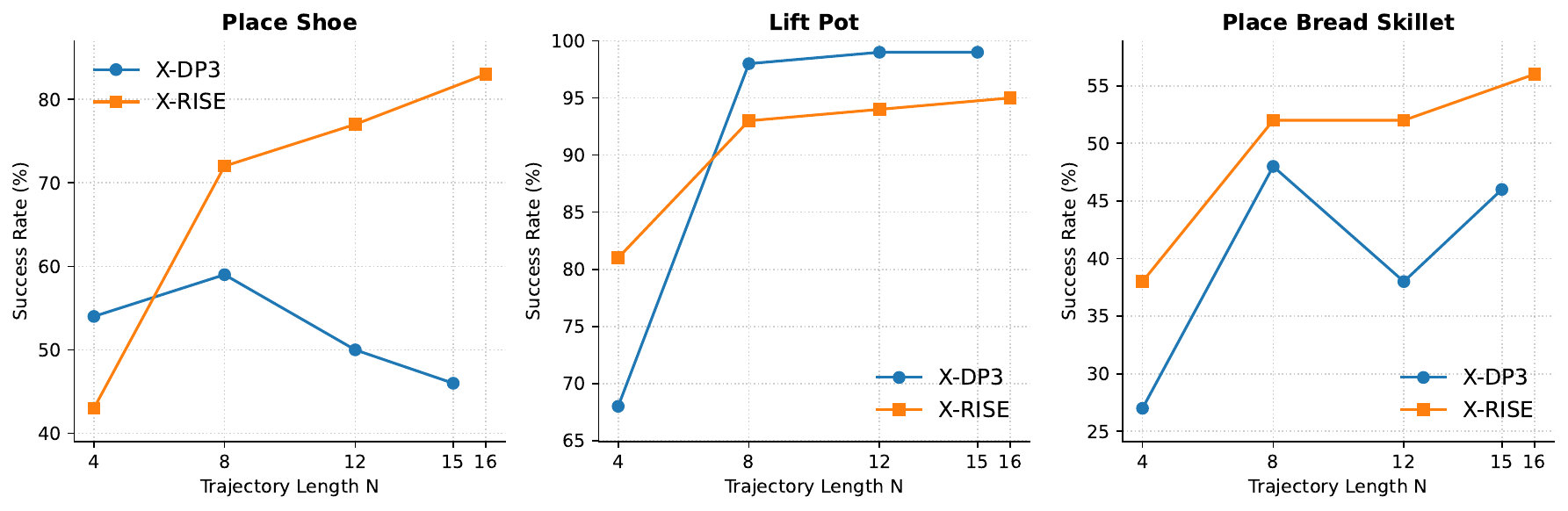}
\caption{\label{fig:line_chart} Ablation study of the conditional trajectory length $N$ which defaults to 8 in all experiments.}
\vspace{-0.5cm}
\end{figure}

To study the effect of trajectory length, we vary $N$ for X-DP3 and X-RISE on three randomly selected RoboTwin 2.0 tasks. Note that X-DP3 follows the convention of DP3 and executes at most $H - n + 1$ steps, where $n$ denotes the number of observation frames. As shown in Fig.~\ref{fig:line_chart}, X-RISE exhibits a clear performance improvement as $N$ increases, whereas X-DP3 saturates at $N = 8$. This may stem from the limited capacity of DP3, as discussed in~\cite{wang2024rise}. Both methods experience a notable performance drop at $N = 4$, likely because shorter sequences fail to capture meaningful motion patterns and thus degrade the base policy.

\section{Real-world Experiments}

In this section, we aim to answer the following research question.

\begin{enumerate}[leftmargin=1.5cm]
    \item[\textbf{(RQ4)}] Does the advantage of X-Imitator demonstrated in simulation environments translate to real-world scenarios?
\end{enumerate}

\subsection{Setup}
\subsubsection{Platform.} Our robot platform consists of a Flexiv Rizon 4 robotic arm with a Dahuan AG-95 gripper. A global Intel RealSense D435 RGB-D camera is positioned in front of the robot, perceiving the robot workspace during manipulation. Following~\cite{rh20t}, we use sigma 7 haptic device to teleoperate the robot to gather 50 expert demonstrations for each task.

\begin{figure}
\centering
\includegraphics[width=1.0\linewidth]{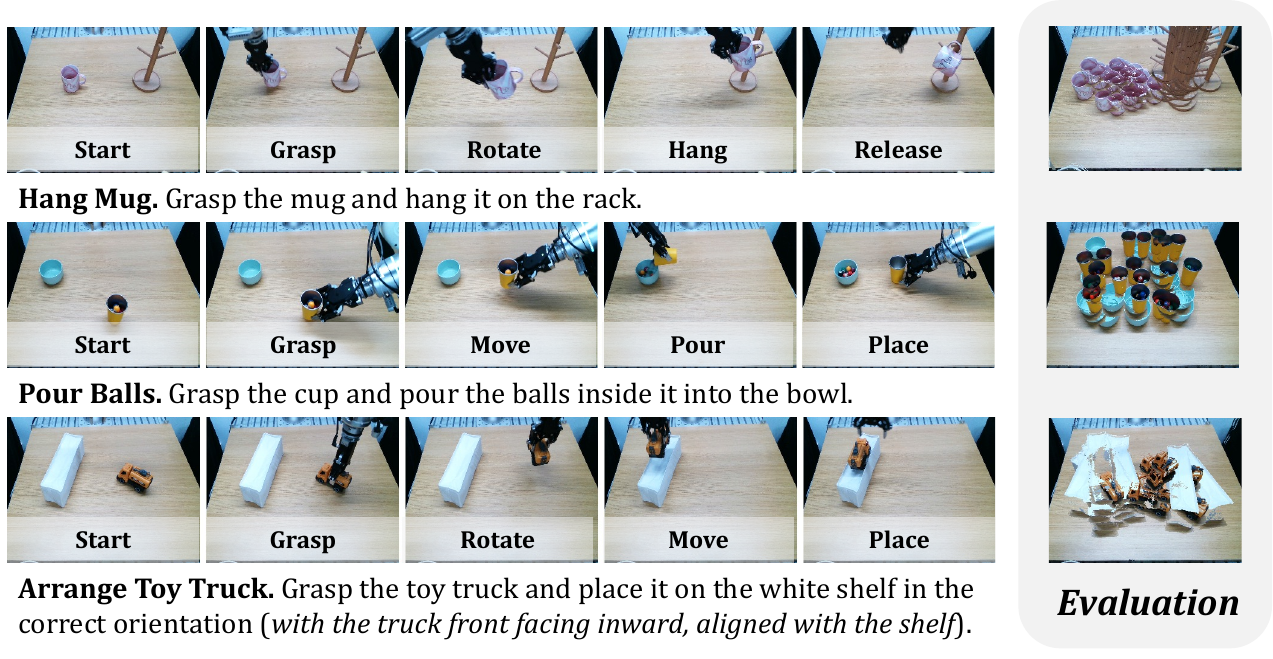}
\caption{\label{fig:real_tasks} Real-world task descriptions. The last column shows the evaluation setup.}
\vspace{-0.5cm}
\end{figure}

\subsubsection{Tasks.} As illustrated in Fig. \ref{fig:real_tasks}, we design three tasks \textit{Hang Mug}, \textit{Pour Balls} and \textit{Pick Place Truck} for the real-world experiments. These tasks demand both fine-grained orientation control and precise positioning.
For instance, in the \textit{Arrange Toy Truck} task, the wheels will roll if the gripper is not aligned properly with the truck body during grasping, and the truck will drop if it is not rotated to align with the narrow white stage.

\subsubsection{Implementations.} After collecting expert demonstrations,  FoundationPose~\cite{wen2024foundationpose} is used to 
estimate the object pose in the first episode frame and to track the remaining ones.
For the inputs to FoundationPose, we utilize text-prompted SAM2~\cite{lang-sam2,ravi2024sam2} to generate an initial object mask, and use the MagiScan app to acquire an up-to-scale 3D mesh which is then manually adjusted to the real-world size within several minutes.
We choose the best-performing visuomotor policy in the simulation experiments, RISE and X-RISE, for real-world evaluations. 

\subsubsection{Evaluation Protocols.}
Following the procedure in~\cite{umi,cage,rise2}, we utilize a consistent evaluation protocol for all policies to minimize performance variance and ensure reproducibility. Specifically, uniformly distributed test positions are randomly generated prior to each task evaluation, as illustrated in the last column of Fig \ref{fig:real_tasks}. Each policy is evaluated for 20 consecutive trials to compute its success rate. We limit the total number of execution steps to 300. To conduct a more detailed comparison, we divide each task into two or several successive stages and record their success rates respectively.

\subsection{Results}
\subsubsection{The X-Imitator framework is robust to real-world sensor noise and physical uncertainties, maintaining its lead in the real-world tasks (RQ4).}
The real-world results summarized in Table \ref{tab:real_result} indicate that X-RISE significantly improves system reliability. 
In the \textit{Hang Mug} task, X-RISE increases the final ``\textit{hang}'' success rate from 60\% to 80\%. 
In the \textit{Pour Balls} task, X-RISE successfully pours a greater number of balls with lower standard deviation. We note that the baseline RISE occasionally fails to position the cup accurately above the bowl, causing many balls to roll off the table. As a result, the policy perceives few or no balls in the bowl yet continues pouring attempts, even when the cup is empty, until the maximum number of execution steps is reached. 
In the \textit{Arrange Toy Truck} task, we observe that RISE sometimes fails to align the gripper properly with the truck body, leading to a lower \textit{grasp} success rate. Both RISE and our X-RISE occasionally place the truck in the wrong direction (facing outward). This misorientation may occur because neither policy utilizes RGB input during our training, which makes it difficult to discern the truck's head orientation from partial point clouds alone.

\begin{table*}[t]
\caption{Evaluation results on real-world tasks. The \#balls column denotes the number of balls successfully poured into the bowl.}
\label{tab:real_result}
\centering
\footnotesize
\setlength{\tabcolsep}{4pt}
\begin{tabular}{lcccccccc}
\toprule
\multirow{2}{*}{\textbf{Method}} & \multicolumn{2}{c}{\textbf{Hang Mug}} & \multicolumn{3}{c}{\textbf{Pour Balls}} & \multicolumn{3}{c}{\textbf{Arrange Toy Truck}} \\
\cmidrule(lr){2-3} \cmidrule(lr){4-6} \cmidrule(lr){7-9} 
 & grasp & hang & grasp & place & \#balls & grasp & place & direction \\

\midrule
RISE~\cite{wang2024rise} & 90\% & 60\% & 90\% & 80\% & 7.4 $\pm$ 3.3 & 65\% & 55\% & 55\% \\
X-RISE \textit{(ours)} & \cellcolor[HTML]{ecf8f8}\textbf{100}\% & \cellcolor[HTML]{ecf8f8}\textbf{80}\% & \cellcolor[HTML]{ecf8f8}\textbf{100}\% & \cellcolor[HTML]{ecf8f8}\textbf{100}\% & \cellcolor[HTML]{ecf8f8}$\textbf{8.9 $\pm$ 1.2}$ & \cellcolor[HTML]{ecf8f8}\textbf{100}\% & \cellcolor[HTML]{ecf8f8}\textbf{80}\% & \cellcolor[HTML]{ecf8f8}\textbf{80}\% \\
\bottomrule
\end{tabular}
\end{table*}

\section{Limitations and Conclusion}

\subsubsection{Limitations.} 
First, the current framework assumes that the manipulated objects can be represented by a rigid SE(3) pose. Thus, our method can not be directly applied to deformable objects (\textit{e.g.}, ropes, cloth), and is limitedly applied to articulated objects when we treat the manipulated part as an individual object.  
Second, the external pose estimator used for obtaining training pose labels may fail under severe occlusion, motion blur, or poor lighting in real-world tasks. Therefore, a correction procedure is required sometimes. 
Third, the framework assumes a structured scene where the target object is known a priori. Handling unstructured scenes remains an open challenge.

\subsubsection{Conclusion.} 
We have introduced X-Imitator, a flexible dual-path imitation learning framework that enhances spatial awareness in robot manipulation via bidirectional action-pose interaction. By reciprocally conditioning current pose predictions on past actions and vice versa, X-Imitator enables mutual refinement of action generation and spatial reasoning. With minimal inference overhead and broad architectural compatibility, X-Imitator offers a simple yet powerful plug-in for existing imitation learning methods with effective spatial perception.


\appendix
\section*{Supplement} 

\section{Domain Randomization}

We generate domain randomized data (clutter, table heights, \etc) to study its impact on the model performance in Table \ref{tab:domain_random}. 
As can be see, X-RISE demonstrates superior performance compared to the base policy RISE, highlighting the benefits of its bidirectional interaction mechanism.
However, both policies show a significant performance drop relative to the results under clean configurations reported in the main paper. This indicates that there remains considerable room for improving imitation learning in handling cluttered scenes.
Notably, X-RISE maintains or even improves the success rate on the \textit{Lift Pot} (pot) task. We speculate that this is because the pot's point cloud has a relatively large size, which helps the model be more robust to distractors. Another somewhat counterintuitive result is that the evaluation results on random data are better than those on clean data. This is because the model was trained on random data and has hardly ever encountered clean data. This reminds us that when it comes to data-driven methods, it is crucial to carefully prepare the training data.

\begin{table*}
\caption{Simulation success rates (\%) on RoboTwin 2.0 benchmark. Both models are trained on domain randomized data while evaluated on clean or randomized data.}
\label{tab:domain_random}
\centering
\footnotesize
\setlength{\tabcolsep}{4pt}
\begin{tabular}{lcccccccc}
\toprule
\multirow{2}{*}{\textbf{Method}} & \multicolumn{3}{c}{\textbf{Single-Arm}} & \multicolumn{4}{c}{\textbf{Dual-Arm}} & \multirow{2}{*}{\textbf{Average}} \\
\cmidrule(lr){2-4} \cmidrule(lr){5-8} 
 & hammer & shoe & QRcode & mug & pot & bread & shoes \\ 
\midrule
RISE \textit{(clean)} & 32 & 9 & 4 & 9 & 83 & 5 & 0 & 20.29 \\
X-RISE \textit{(clean)} & \textbf{42} & \textbf{12} & \textbf{10} & \textbf{13} & \textbf{93} & \textbf{6} & 0 & \textbf{25.14} \\
\midrule
RISE \textit{(random)} & \textbf{41} & 28 & \textbf{53} & 12 & 78 & 7 & 4 & 31.86 \\
X-RISE \textit{(random)} & 40 & \textbf{30} & 46 & \textbf{19} & \textbf{95} & 7 & 4 & \textbf{34.43} \\
\bottomrule
\end{tabular}
\vspace{-\baselineskip}
\end{table*}

\section{Robustness to Historical Prediction Noise}

In our code implementation, we perform condition augmentation by adding noise to the ground-truth action/pose in the training stage, mimicking prediction error during inference. Fig. \ref{fig:line_chart} of the main paper shows that performance improves as the conditional length $N$ increases, even as prediction noise also increases alongside useful historical information. It indicates that the framework has anti-noise ability and can effectively extract valuable patterns from history.

\begin{figure}
    \centering
    \begin{subfigure}[b]{0.48\linewidth}
        \centering
        \includegraphics[width=\linewidth]{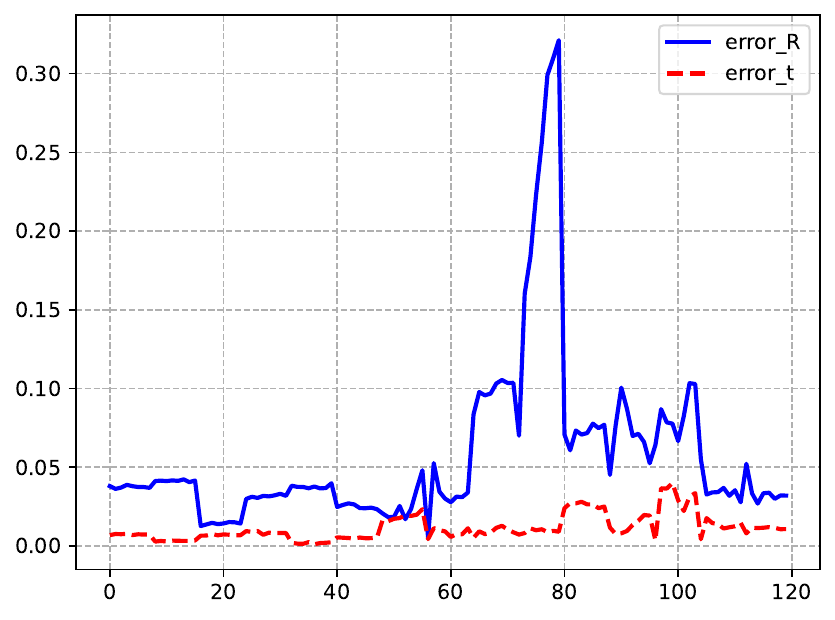}
        \caption{Episode 1: Success}
    \end{subfigure}
    \hfill
    \begin{subfigure}[b]{0.48\linewidth}
        \centering
        \includegraphics[width=\linewidth]{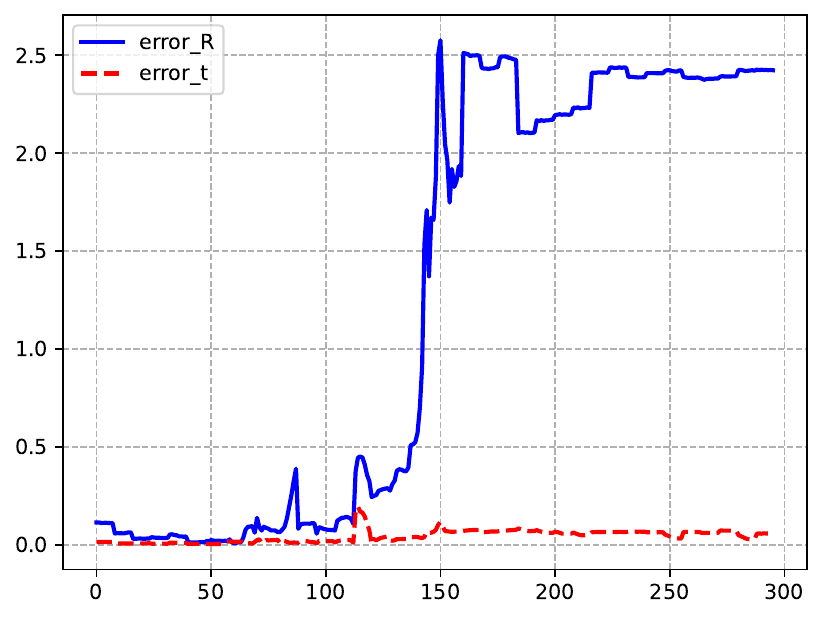}
        \caption{Episode 2: Failure}
    \end{subfigure}
    \begin{subfigure}[b]{0.48\linewidth}
        \centering
        \includegraphics[width=\linewidth]{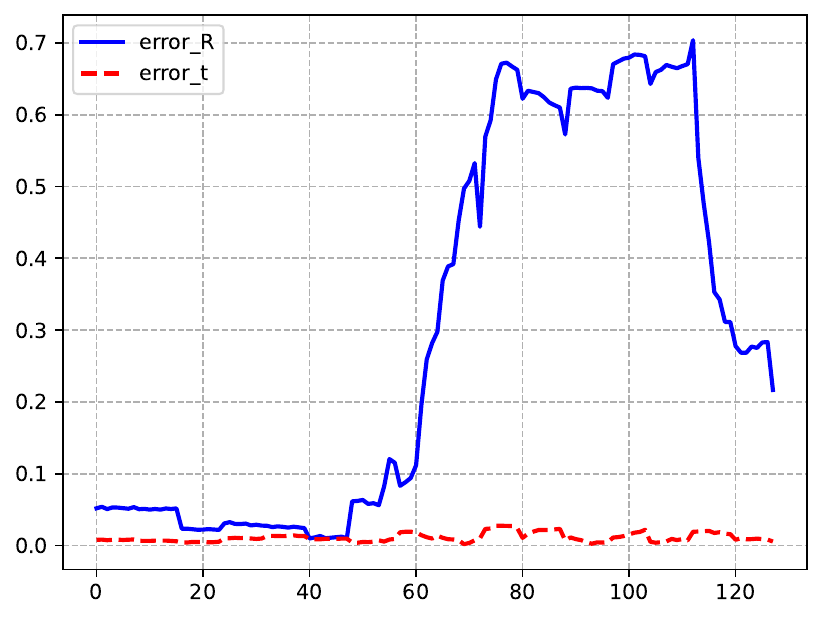}
        \caption{Episode 3: Success}
    \end{subfigure}
    \hfill
    \begin{subfigure}[b]{0.48\linewidth}
        \centering
        \includegraphics[width=\linewidth]{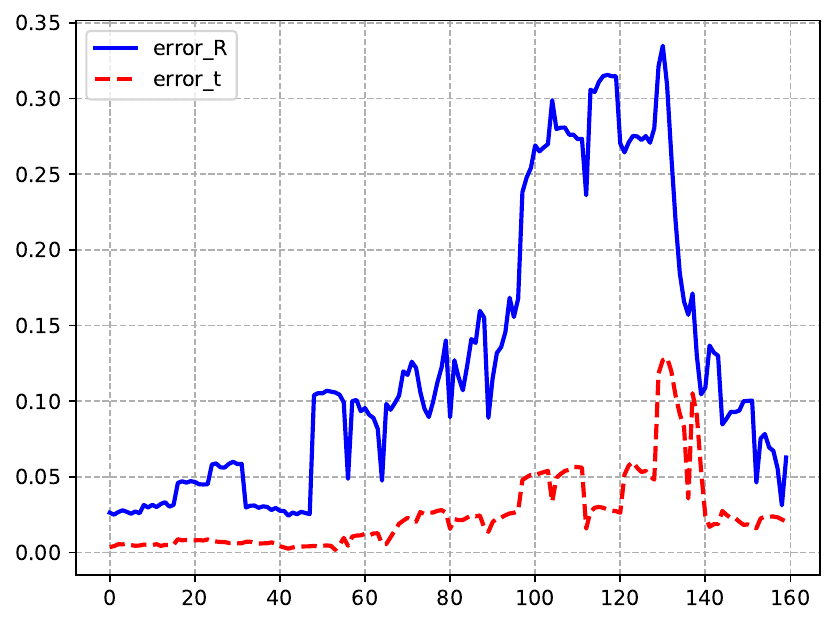}
        \caption{Episode 4: Success}
    \end{subfigure}
    
    \caption{Per-step error curves during inference for the \textit{Rotate QRcode} task.}
    \label{fig:per_step_error}

\vspace{-\baselineskip}
\end{figure}

Table \ref{tab:oracle_test} further demonstrates robustness to historical prediction noise. When the standard deviation of the noise added to \( (R,t) \) is set to (0.2 rad, 0.02 m), X-RISE and its (gt+noise) version yield close results. It suggests that the most prediction errors of X-RISE also fall within (34\degree, 6cm) under the $3\sigma$ principle. We plot the per-step error curves to verify this, and find that an occasional error increase can usually be corrected in the subsequent steps, as illustrated in Fig.~\ref{fig:per_step_error}.

\begin{table}
\caption{Oracle test on RoboTwin 2.0 benchmark. (gt) uses the ground-truth conditional pose trajectory. (gt+noise) adds noise from a zero-mean normal distribution. There are no ground-truth actions during inference.}
\label{tab:oracle_test}
\centering
\footnotesize
\setlength{\tabcolsep}{4pt}
\begin{tabular}{lccccc}
\toprule
\textbf{Method} & \textbf{hammer} & \textbf{shoe} & \textbf{QRcode} & \textbf{Average} \\ 
\midrule
X-RISE              & 92 & 72 & 63 & 75.7 \\
X-RISE(gt)          & 93 & 79 & 66 & 79.3 \\
X-RISE(gt+noise)    & 91 & 73 & 62 & 75.3 \\
\bottomrule
\end{tabular}
\vspace{-\baselineskip}
\end{table}

\section{Label Pipeline for Real-World Tasks}

\begin{itemize}
    \item \textbf{Mask}:
    We only need to segment the first frame of each episode by text-prompted SAM2 (see Section 5.1 for reference) which can be upgraded to SAM3.

    \item \textbf{Mesh}:
    The "One Image to 3D" function of MagiScan is enough for us to obtain a good up-to-scale 3D mesh. We then manually adjust to its real size. This costs several minutes and is scalable.

    \item \textbf{Correction}:
    Occurred in 35 out of 150 episodes of the three real-world tasks, normal motion blur and partial occlusion caused large track errors within a short frame sequence. We reran text-prompted SAM2 and FoundationPose to estimate the first failed frame and to track the remaining ones. Just one time correction solved the issue. Humans only perform inspection and the cost is acceptable.
    
\end{itemize}

\bibliographystyle{splncs04}
\bibliography{main}
\end{document}